\documentclass[11pt]{article}

\usepackage{jmlr2e}
\usepackage[utf8]{inputenc}
\usepackage[T1]{fontenc}
\usepackage[mathscr]{euscript}
\usepackage{amsmath}
\usepackage{bm}
\usepackage{newtxmath}

\usepackage{subcaption}
\usepackage{graphicx}
\usepackage{multirow}
\usepackage{booktabs}
\usepackage{tabularx}
\newcolumntype{Y}{>{\centering\arraybackslash}X}

\ShortHeadings{}{}
\firstpageno{1}

\begin{document}

\title{Nonlinear Granger Causality using Kernel Ridge Regression}

\author{\name Wojciech ``Victor'' Fulmyk \email fulmykw@gmail.com}

\maketitle

\vspace{0.5cm}





\begin{abstract}
I introduce a novel algorithm and accompanying Python library, named \href{https://github.com/WojtekFulmyk/mlcausality}{\textbf{mlcausality}}, designed for the identification of nonlinear Granger causal relationships. This novel algorithm uses a flexible plug-in architecture that enables researchers to employ any nonlinear regressor as the base prediction model. Subsequently, I conduct a comprehensive performance analysis of \href{https://github.com/WojtekFulmyk/mlcausality}{\textbf{mlcausality}} when the prediction regressor is the kernel ridge regressor with the radial basis function kernel. The results demonstrate that \href{https://github.com/WojtekFulmyk/mlcausality}{\textbf{mlcausality}} employing kernel ridge regression achieves competitive AUC scores across a diverse set of simulated data. Furthermore, \href{https://github.com/WojtekFulmyk/mlcausality}{\textbf{mlcausality}} with kernel ridge regression yields more finely calibrated $p$-values in comparison to rival algorithms. This enhancement enables \href{https://github.com/WojtekFulmyk/mlcausality}{\textbf{mlcausality}} to attain superior accuracy scores when using intuitive $p$-value-based thresholding criteria. Finally, \href{https://github.com/WojtekFulmyk/mlcausality}{\textbf{mlcausality}} with kernel ridge regression exhibits significantly reduced computation times compared to existing nonlinear Granger causality algorithms. In fact, in numerous instances, this innovative approach achieves superior solutions within computational timeframes that are an order of magnitude shorter than those required by competing algorithms.

\end{abstract}




\section{Introduction}

Identifying the direction of relationships amongst simultaneously observed time-series continues to generate significant interest among researchers from many different fields. Since Nobel prize-winning economist Clive Granger first introduced the concept of Granger causality \citep{granger1969investigating}, significant theoretical contributions towards the development of Granger-causal methods have been made by computer scientists, biostatisticians, physicists, mathematicians, and many others. Much of the recent research has focused on identifying nonlinear Granger causality, possibly in the presence of other confounding time-series \citep{wismuller2021large,rosol2022granger,lacasa2015network,gao2017complex,dsouza2017exploring}. The novel approach proposed herein, which I call \href{https://github.com/WojtekFulmyk/mlcausality}{\textbf{mlcausality}}, tackles the problem of identifying nonlinear relationships using a regressor-agnostic non-parametric test. Consequently, \href{https://github.com/WojtekFulmyk/mlcausality}{\textbf{mlcausality}} has a plug-in architecture that allows the researcher to use any nonlinear regressor, such as a kernel ridge regressor, a support vector regressor, a random forest regressor, or a gradient boosting regressor, as the base prediction model. Due to \href{https://github.com/WojtekFulmyk/mlcausality}{\textbf{mlcausality}}'s plug-in architecture and the large number of regressors with which it can be used, the rest of this paper focuses exclusively on analyzing \href{https://github.com/WojtekFulmyk/mlcausality}{\textbf{mlcausality}}'s performance when the kernel ridge regressor with the radial basis function kernel is used, with an analysis of the performance of other regressors deferred to subsequent papers. For many nonlinear networks \href{https://github.com/WojtekFulmyk/mlcausality}{\textbf{mlcausality}} with kernel ridge regression outperforms rival algorithms in terms of both network recovery performance as well as execution speed.

\section{Methodology}
\subsection{An introduction to Granger causality}

Fundamentally, Granger causality describes predictability: given two time-series $X$ and $Y$, if the histories of both $X$ and $Y$ predict future values of $Y$ better than the history of $Y$ alone, then $X$ is said to Granger cause $Y$. Before continuing, it is important to note that Granger causality itself is a misnomer because no truly causal or mechanical link between the time-series in question is implied. In other words, if $X$ Granger causes $Y$, then it could be the case that $X$ actually causes $Y$, or it could be the case that $X$ simply leads $Y$ and therefore provides useful information about the future value(s) of $Y$ without actually causing $Y$.

Classical Granger causality is formulated as follows. Given two time-series $X$ and $Y$ and all available information $U$, $X$ Granger causes $Y$ if the variance of the prediction error when the history of $X$ is included is lower than the variance of the prediction error when $X$ is excluded:
\begin{equation}\label{eq:Granger_var_pred_error}
\sigma^2(Y|U) < \sigma^2(Y|(U-X))
\end{equation}
where $(U-X)$ denotes all available information except for time series $X$.

To evaluate equation \ref{eq:Granger_var_pred_error} two linear models on lagged variables are estimated, an ``unrestricted'' model that includes the lags of $X$ and a ``restricted'' model that excludes the lags of $X$. Although it is theoretically possible to include different amounts of lags for each time-series, in practice an equal amount of lags for all time-series is typically used. Given an equal amount of lags for all variables, the classical Granger causality model evaluates:
\begin{align}
&Y(t) = \alpha_{r,0} + \sum_{l=1}^L{\alpha_{r,l}Y(t-l)} + E_r \label{eq:Granger_classic_r}\\
&Y(t) = \alpha_{u,0} + \sum_{l=1}^L{\alpha_{u,l}Y(t-l)} + \sum_{l=1}^L{\beta_{u,l}X(t-l)} + E_u \label{eq:Granger_classic_u}
\end{align}

\noindent where $L$ is the total number of lags; $\alpha_{m,0}$ are the intercepts for the restricted $r$ and unrestricted $u$ models $m$; $\alpha_{m,l}$ are the coefficients for lags $l$ of $Y$; $\beta_{m,l}$ are the coefficients for lags $l$ of $X$, and $E_m$ are the error terms.

Note that, under classical Granger causality, the restricted model is a nested model of the unrestricted model. For linear regression this guarantees that the fit of the unrestricted model will be no worse than that of the restricted model. In order to test whether the variance of the unrestricted model is significantly lower, in the statistical sense, than the variance of the restricted model, an F-test is performed:
\begin{equation}\label{eq:F-test}
F = \frac{RSS_r- RSS_u/L}{RSS_u/(N-2L-1)}
\end{equation}
where $N$ is the total number of training samples for which all lag terms exist, $RSS_m$ is the residual sum of squares for model $m$, and $L$ is the total number of lags. Under the null hypothesis, if the fit from the unrestricted model is not significantly better than that of the restricted model, the $F$-statistic from equation \ref{eq:F-test} follows an $F$-distribution with $(L, N-2L-1)$ degrees of freedom.

The primary limitation of the classical Granger causality model is that, by construction, the classical model is only able to identify Granger causality if $X$ and $Y$ are linearly related. Identifying nonlinear relationships using classical Granger causality would require performing nonlinear data transformations, such as by taking natural logs of the data. However, making such explicit data transformations is not always a trivial task, especially in cases where the data generating process is not well known to the researcher.

On a positive note, the addition of superfluous confounding time-series into the classical model is possible as long as the denominator degrees of freedom remain strictly positive. In other words, if there was an additional confounding time-series $Z$ that one would want to include in the analysis, then one could simply add $\sum_{l=1}^L{\gamma_{m,l}Z(t-l)}$ to both the restricted and unrestricted models in equations \ref{eq:Granger_classic_r} and \ref{eq:Granger_classic_u} and alter the denominator degrees of freedom in equation \ref{eq:F-test} to $N-3L-1$. The null hypothesis with confounding time-series $Z$ is that $\sigma^2(Y|X_{t-1},...,X_{t-L}, Y_{t-1},...,Y_{t-L}, Z_{t-1},...,Z_{t-L}) \nless \sigma^2(Y|Y_{t-1},...,Y_{t-L}, Z_{t-1},...,Z_{t-L})$. Additional confounding time-series can be added analogously.

\subsection{Kernel Ridge Regression}
The following provides a brief review of kernel ridge regression. For a more thorough treatment, the reader is encouraged to review \cite{murphy2012machine,vovk2013kernel}; and \cite{exterkate2016nonlinear}.

Let $\boldsymbol{x} \in \mathbb{R}^D$ represent a feature vector for $D$ features, and $\boldsymbol{X} \in \mathbb{R}^{N \times D}$ represent a design matrix for $N$ observations. Given a target vector $\boldsymbol{y} \in \mathbb{R}^N$, ridge regression minimizes the following objective function:
\begin{equation}\label{eq:ridge-reg}
\begin{aligned}
&\min_{w} \ || \boldsymbol{y} - \boldsymbol{X} \boldsymbol{w} ||_2 ^2 + \lambda || \boldsymbol{w} ||_2 ^2 \\
= \ &\min_{w} \ (\boldsymbol{y} - \boldsymbol{X} \boldsymbol{w})^T (\boldsymbol{y} - \boldsymbol{X} \boldsymbol{w}) + \lambda \boldsymbol{w}^T \boldsymbol{w}
\end{aligned}
\end{equation}
where $\boldsymbol{w} \in \mathbb{R}^D$ are the regression coefficients and $\lambda$ is a penalty term that penalizes the magnitudes of those coefficients. Note that $\lambda$ acts as a regularization term: if $\lambda = 0$, then objective function \ref{eq:ridge-reg} collapses to a linear regression, while values of $\lambda$ greater than zero encourage smaller coefficient magnitudes at the optimum.

The solution is found by taking the derivative of the objective function and setting that derivative equal to zero:
\begin{align}
\boldsymbol{w} = (\boldsymbol{X}^T \boldsymbol{X} + \lambda \boldsymbol{I}_D)^{-1}\boldsymbol{X}^T \boldsymbol{y} \label{eq:ridge-reg-solution1}\\
\boldsymbol{w} = \boldsymbol{X}^T(\boldsymbol{X} \boldsymbol{X}^T + \lambda \boldsymbol{I}_N)^{-1} \boldsymbol{y} \label{eq:ridge-reg-solution2}
\end{align}
where equation \ref{eq:ridge-reg-solution2} is due to the Sherman-Morrison-Woodbury formula \citep{woodbury1950inverting}.

The ridge regression model, as described above, produces nothing more than a linear solution that penalizes the magnitudes of the coefficients in relation to $\lambda$ and is therefore incapable of efficiently modelling nonlinear relationships by itself. In order to introduce nonlinearity, the ridge regression model is kernalized. Intuitively, kernel methods map feature vectors into a higher, possibly infinitely dimensional space. The goal of kernel ridge regression is to then apply (inherently linear) ridge regression on the mappings of the feature vectors in that higher dimensional space. However, from a computational perspective, it turns out that one does not need to actually map feature vectors into a different space at all; rather, it is sufficient to just calculate relationships between all feature vectors using a kernel function. This process is commonly referred to as the ``kernel trick.''

Formally, define a symmetric kernel function $\kappa(\boldsymbol{x},\boldsymbol{x}') = \kappa(\boldsymbol{x}',\boldsymbol{x}) \in \mathbb{R}$ as a real-valued function for feature vectors $\boldsymbol{x},\boldsymbol{x}' \in \chi$, with $|\chi| = N$. If the Gram matrix $\boldsymbol{K}$ with elements $\kappa(\boldsymbol{x}_i,\boldsymbol{x}_j)_{ij} \ \forall \ i,j \in \{1,...,N\}$  is positive definite, then by Mercer's theorem \citep{mercer1909xvi} there exists a function $\phi$ mapping $\boldsymbol{x}$ to $\mathbb{R}^M$ such that 
\begin{equation}\label{eq:mercer-theorem}
\kappa(\boldsymbol{x},\boldsymbol{x}') = \phi(\boldsymbol{x})^T \phi(\boldsymbol{x}')
\end{equation}

Moreover, Mercer's theorem guarantees that $M$, the dimension of the space to which $\phi$ maps feature vectors into, can be arbitrarily large, potentially infinite. A kernel that satisfies the above Gram matrix condition is called a Mercer kernel.

Ridge regression can be kernalized as follows. Suppose that one wishes to evaluate ridge regression not on the feature vectors $\boldsymbol{x}$ directly but rather on some mapping $\phi(\boldsymbol{x})$ into a higher, possibly infinitely dimensional, space. Moreover, let $\boldsymbol{\phi}(\cdot)$ represent the multi-observational analogue of $\phi(\cdot)$ that admits $\boldsymbol{X}$, the design matrix that stores feature vectors $\boldsymbol{x}$ for multiple observations, as an input. Then ridge regression could be partially kernalized by replacing $\boldsymbol{X} \boldsymbol{X}^T$ in equation \ref{eq:ridge-reg-solution2} with the Gram matrix $\boldsymbol{K}$ for some Mercer kernel $\kappa(\boldsymbol{x},\boldsymbol{x}') = \phi(\boldsymbol{x})^T \phi(\boldsymbol{x}')$, and by replacing $\boldsymbol{X}^T$ with $\boldsymbol{\phi} (\boldsymbol{X})^T$. The partially kernalized solution then becomes:
\begin{align}
\boldsymbol{w} = \boldsymbol{\phi} (\boldsymbol{X})^T(\boldsymbol{K} + \lambda \boldsymbol{I}_N)^{-1} \boldsymbol{y} \label{eq:ridge-reg-solution3}
\end{align}
Now, let $\boldsymbol{\alpha} = (\boldsymbol{K} + \lambda \boldsymbol{I}_N)^{-1} \boldsymbol{y}$. Equation \ref{eq:ridge-reg-solution3} then becomes:
\begin{align}
\boldsymbol{w} &= \boldsymbol{\phi} (\boldsymbol{X})^T \boldsymbol{\alpha} \label{eq:ridge-reg-solution4} \\
&= \sum_{i=1}^N \alpha_i \phi(\boldsymbol{x}_i) \label{eq:ridge-reg-solution5}
\end{align}

Note that equation \ref{eq:ridge-reg-solution4} is not yet usable because the kernalization is only partial: we still have to deal with $\boldsymbol{\phi} (\boldsymbol{X})^T$, which is potentially difficult to evaluate because all feature vectors in $\boldsymbol{X}$ would have to be transformed using $\phi(\cdot)$ directly. However, if just the in-sample predictions from the kernel ridge regression are needed, then for each feature vector $\boldsymbol{x}$, one just has to evaluate
\begin{align}
\hat{f}(\boldsymbol{x}) &= \boldsymbol{w}^T \phi(\boldsymbol{x}) \label{eq:ridge-reg-solution6} \\
&= \sum_{i=1}^N \alpha_i \phi(\boldsymbol{x}_i)^T \phi(\boldsymbol{x}) \label{eq:ridge-reg-solution7} \\
&= \sum_{i=1}^N \alpha_i \kappa(\boldsymbol{x}_i,\boldsymbol{x}) \label{eq:ridge-reg-solution8}
\end{align}
where the last line completes the kernalization. It is thus possible to generate nonlinear in-sample predictions by fitting an (inherently linear) ridge regression on a mapping of the feature vectors into a higher dimensional space, without actually performing the mapping into that higher dimensional space. In order to perform out-of-sample predictions, one would first fit equation \ref{eq:ridge-reg-solution8} using the training data and then use the obtained in-sample predictions to precompute $\boldsymbol{\alpha}$; out-of-sample predictions could then be made directly using equation \ref{eq:ridge-reg-solution8} with that precomputed $\boldsymbol{\alpha}$ and a previously unseen feature vector $\boldsymbol{x}$ that comes from the test set.

Using kernel ridge regression in practice involves choosing a suitable kernel function and, by design, the \href{https://github.com/WojtekFulmyk/mlcausality}{\textbf{mlcausality}} Python library does not provide any restrictions with regards to this: the user is free to choose any kernel function supported by the \textbf{scikit-learn} Python library. However, all of the subsequent analysis presented herein was performed using the radial basis function (RBF) kernel, also known as the Gaussian kernel:
\begin{align}\label{eq:rbf}
\kappa(\boldsymbol{x},\boldsymbol{x}') = e^{-\gamma ||\boldsymbol{x} - \boldsymbol{x}'||_2^2}
\end{align}
where $\gamma \in (0,\infty)$ is a hyperparameter. The RBF kernel is infinitely dimensional, as can be seen in the multinomial expansion in \cite{shashua2009introduction}.

\subsection{The \href{https://github.com/WojtekFulmyk/mlcausality}{\textbf{mlcausality}} Algorithm}

Suppose there is a set $\vmathbb{X}$ of equally-spaced time-series that share a common time index and do not contain any gaps. Suppose that each of these time series is represented by a column vector $\vmathbb{x} \in \vmathbb{X}$ and that each of these vectors has length $N+L$. Suppose that $\vmathbb{X}$ contains at least 2 such time series, that is, that $|\vmathbb{X}| = G \geq 2$. Let $L$ represent the number of lags. Moreover, let $\boldsymbol{X}_u \in \mathbb{R}^{N \times LG}$ be a matrix of $L$ lags $\forall$ $G$ time-series in $\vmathbb{X}$, as depicted in the matrix below:

\[
\mathbf{\boldsymbol{X}_u} = \begin{bmatrix}
    \vmathbb{x}_{1,1\phantom{-1}} & \cdots & \vmathbb{x}_{1,L\phantom{-1}\phantom{-1}} & \cdots \cdots & \vmathbb{x}_{G,1\phantom{-1}} & \cdots & \vmathbb{x}_{G,L\phantom{-1}\phantom{-1}} \\
    \vmathbb{x}_{1,2\phantom{-1}} & \cdots & \vmathbb{x}_{1,L+1\phantom{-1}} & \cdots \cdots & \vmathbb{x}_{G,2\phantom{-1}} & \cdots & \vmathbb{x}_{G,L+1\phantom{-1}} \\
    \vmathbb{x}_{1,3\phantom{-1}} & \cdots & \vmathbb{x}_{1,L+2\phantom{-1}} & \cdots \cdots & \vmathbb{x}_{G,3\phantom{-1}} & \cdots & \vmathbb{x}_{G,L+2\phantom{-1}} \\
    \vdots & \cdots & \vdots & \ddots & \vdots & \cdots & \vdots \\
    \vmathbb{x}_{1,N-2} & \cdots & \vmathbb{x}_{1,N+L-3} & \cdots \cdots & \vmathbb{x}_{G,N-2} & \cdots & \vmathbb{x}_{G,N+L-3} \\
    \vmathbb{x}_{1,N-1} & \cdots & \vmathbb{x}_{1,N+L-2} & \cdots \cdots & \vmathbb{x}_{G,N-1} & \cdots & \vmathbb{x}_{G,N+L-2} \\
    \vmathbb{x}_{1,N\phantom{-1}} & \cdots & \vmathbb{x}_{1,N+L-1} & \cdots \cdots & \vmathbb{x}_{G,N\phantom{-1}} & \cdots & \vmathbb{x}_{G,N+L-1} \\
\end{bmatrix}
\]
where the first coordinate in the subscript of an element indicates the time-series and the second coordinate represents the time index. Note that a single row in the above matrix represents a concatenation of ${1,...,L}$ lags $\forall$ $G$ time-series in $\vmathbb{X}$, and that only rows that have no missing lags are kept. Similarly, let $\boldsymbol{X}_r \in \mathbb{R}^{N \times L(G-1)}$ be a matrix of $L$ lags $\forall$ $G-1$ time-series in $\vmathbb{X}$ other than time-series $\boldsymbol{q} \in \vmathbb{X}$. Then, in order to test whether time-series $\boldsymbol{q} \in \vmathbb{X}$ Granger-causes time-series $\boldsymbol{y} \in \vmathbb{X}$ with $\boldsymbol{q} \neq \boldsymbol{y}$, \href{https://github.com/WojtekFulmyk/mlcausality}{\textbf{mlcausality}} evaluates
\begin{align}
&\hat{\boldsymbol{y}}_r = f_r (\boldsymbol{X}_r, \boldsymbol{y}_{true},*_r) \label{eq:mlcasality-r} \\
&\hat{\boldsymbol{y}}_u = f_u (\boldsymbol{X}_u, \boldsymbol{y}_{true},*_u) \label{eq:mlcasality-u}
\end{align}
where the subscripts $r$ and $u$ represent the restricted and unrestricted models, respectively; $f$ represents any nonlinear regressor, such as the kernel ridge regressor described above; and $*$ is a placeholder for any additional hyperparameters the chosen regressor accepts.

For kernel ridge regression with the RBF kernel $*_r = (\lambda, \gamma_r)$ and $*_u = (\lambda, \gamma_u)$. Note that, in theory, $\lambda$ can be varied between the restricted and unrestricted models, however, different $\lambda$ values for the restricted and unrestricted models would make it difficult to assess whether the restricted and unrestricted models perform differently solely on the bases of the exclusion or inclusion of time-series $\boldsymbol{q}$; as such, $\lambda$ will be kept the same for both models. It is not clear which value of $\lambda$ should be chosen for the nonlinear Granger causality identification task, although simulation results suggest that the performance of \href{https://github.com/WojtekFulmyk/mlcausality}{\textbf{mlcausality}} with the kernel ridge regressor and the RBF kernel is not very sensitive to the choice of $\lambda$. Consequently, all results presented herein use $\lambda = 1$, which is the \textbf{scikit-learn} Python library default value for the penalty term in kernel ridge regression.

On the other hand, $\gamma$ is subscripted with the model type because it is natural to use $\gamma = 1/D$, where $D$ is the number of features, i.e. the number of columns in either $\boldsymbol{X}_r$ or $\boldsymbol{X}_u$, for the problem at hand. This is because parameter $\gamma$ in the RBF kernel acts as a multiplier for the square of the Euclidean norm, where the number of elements in the sum of that norm is equal to the number of elements in a feature vector. Using an identical $\gamma$ for both models would not account or correct for the greater amount of features in the unrestricted model stemming from its inclusion of time-series $\boldsymbol{q}$. As such, all results presented herein use $\gamma = 1/D$ which acts as a weight that scales the squared norm in the RBF kernel by the number of features in that norm.

\subsubsection{The Statistical Test}

A notable distinction between classical Granger causality and \href{https://github.com/WojtekFulmyk/mlcausality}{\textbf{mlcausality}} lies in the substitution of linear regression with a nonlinear regressor, such as the kernel ridge regressor or the support vector regressor. The task now is to identify an appropriate approach for assessing whether the forecasts generated by the unrestricted model exhibit superior performance compared to those produced by the restricted model in the context of the usage of that nonlinear regressor.

Recall that classical Granger causality relies on an $F$-test to determine if the variance of the unrestricted model significantly differs from that of the restricted model. Nevertheless, the incorporation of nonlinear regressors within the framework of \href{https://github.com/WojtekFulmyk/mlcausality}{\textbf{mlcausality}} can potentially lead to suboptimal outcomes due to the $F$-test's sensitivity to normality, or even to an outright violation of some of the assumptions underpinning the $F$-test. Furthermore, depending on the specific regressor employed, it may prove challenging or even infeasible to precisely calculate the required degrees of freedom for executing the $F$-test. Consequently, instead of resorting to the $F$-test, \href{https://github.com/WojtekFulmyk/mlcausality}{\textbf{mlcausality}} opts for the sign test \citep{dixon1946statistical}, a non-parametric test that imposes minimal assumptions.

The sign test compares the counts of positive and negative values and checks whether they follow a binomial distribution with the probability of success set to 0.5. The null hypothesis of the sign test is that the median of the distribution is equal to zero, while the one-sided alternative hypothesis is that the the median is greater than zero.

The data subjected to the sign test is formulated in the following manner. First, the absolute values of the errors from the restricted and unrestricted models are calculated: 
\begin{align}
& |\boldsymbol{E}_r| = |\hat{\boldsymbol{y}}_r - \boldsymbol{y}_{true}|  \\
& |\boldsymbol{E}_u| = |\hat{\boldsymbol{y}}_u - \boldsymbol{y}_{true}|
\end{align}
where $| \boldsymbol{v} |$ indicates, through an abuse of notation quite common in computer science, the absolute value of every element in vector $\boldsymbol{v}$. Then, a difference between the absolute values of the errors is found: $\boldsymbol{\delta} = |\boldsymbol{E}_r| - |\boldsymbol{E}_u|$. The sign test is subsequently employed on $\boldsymbol{\delta}$.

There are several issues related to the usage of the sign test in this context that must be addressed. First off, note that, although the sign test does not come with very restrictive assumptions, there is one key assumption that must be met: that of independence of all elements in $\boldsymbol{\delta}$. For the context presented herein independence can be assured by splitting the original set of time-series data into distinct train and test sets on the time domain. Once the split is formed, both the restricted and unrestricted models are trained on the train data only, with the difference vector $\boldsymbol{\delta}$ being constructed solely from the restricted and unrestricted models' predictions of the test data. Note that the train-test split has to account for the time-series characteristics of the underlying data; in other words, the training data has to precede the test data, and there has to be a gap of at least $L$ observations between the train and test sets in order to avoid data leakage. In order to satisfy the above, by default, the \href{https://github.com/WojtekFulmyk/mlcausality}{\textbf{mlcausality}} Python library uses the first 70\% of the observations as the training data and the remaining 30\% of the observations as the testing data, with a gap of $L$ observations between these two datasets. All results presented in this paper are with respect to this default data split, although superior results could be obtained with different splits in some cases. Note that the need to split the data into distinct train and test sets represents a departure from classical Granger causality where the train and test sets are identical and equal to all the data that is available. The main implication of this departure is that, in order for the \href{https://github.com/WojtekFulmyk/mlcausality}{\textbf{mlcausality}} test to work as intended, the train and test data have to come from the same distribution.

Secondly, it is important to acknowledge that the sign test allocates identical weights to all time periods and does not take into account the extent to which one model outperforms the other at a specific time point. An alternative to the sign test that assigns more weight to time periods in $|\boldsymbol{\delta}|$ that have greater values is the Wilcoxon signed rank test \citep{wilcoxon1945individual}. The null hypothesis of the Wilcoxon signed rank test is that the difference between two related paired samples is symmetric around a value less than or equal to zero, while the alternative hypothesis is that the difference is symmetric around a value greater than zero. The Wilcoxon signed rank test evaluates the following test statistic:
\begin{align}\label{eq:wilcoxon}
T = \sum_{i=1}^{N} sgn(\delta_i) R_i
\end{align}
where $sgn(\delta_i)$ is the sign of $\delta_i$, and $R_i$ is the rank of the magnitude of $\delta_i$ in vector $\boldsymbol{\delta}$, with the smallest magnitude being assigned rank one, the second smallest magnitude being assigned rank two, and so on.

The Wilcoxon signed rank test would provide greater power than the sign test if the assumptions of the Wilcoxon signed rank test were fully met; in particular, if all elements in $\boldsymbol{\delta}$ were independent, and if the distribution of $\boldsymbol{\delta}$ was symmetric. The independence assumption can be satisfied with a suitable train-test split, but the symmetry assumption required by the Wilcoxon signed rank test is more restrictive than the minimalist assumptions of the sign test, which only requires the satisfaction of independence. Consequently, the \href{https://github.com/WojtekFulmyk/mlcausality}{\textbf{mlcausality}} Python library uses the sign test by default, although there are overrides that allow the usage of the Wilcoxon signed rank test for those that wish to use it. In practice, at least for the simulated networks presented herein, the sign test tends to perform very well, and there is typically no need to depart from the sign test in favour of the Wilcoxon signed rank test.

\subsubsection{Data Preprocessing}

Prior to feeding the raw data into the \href{https://github.com/WojtekFulmyk/mlcausality}{\textbf{mlcausality}} algorithm, as defined above, some preprocessing steps may be needed or desired.

First off, as is typical for these types of analysis, one should ensure that all time-series fed into the \href{https://github.com/WojtekFulmyk/mlcausality}{\textbf{mlcausality}} algorithm are stationary. Moreover, depending on the regressor that is used, additional transformations may be desired. When the kernel ridge regressor is used all time-series should be, at a minimum, scaled to similar magnitudes. This is necessary because, if there are large discrepancies in the magnitudes of the features, then the effect of penalty parameter $\lambda$ will differ substantially from one feature to the next, which could cause some features to either under-penalized or over-penalized compared to others based on their scale alone.

In practice, for the \href{https://github.com/WojtekFulmyk/mlcausality}{\textbf{mlcausality}} algorithm exclusively, the time-series data for analysis presented herein underwent a transformation using a quantile transformer. To be more specific, each individual time-series was partitioned into 1000 quantiles and then subjected to a transformation that mapped them onto a uniform distribution scaled within the range of 0 to 1. This particular transformation yields outstanding results when employed in conjunction with the sign test utilized by \href{https://github.com/WojtekFulmyk/mlcausality}{\textbf{mlcausality}}. With this quantile transformation, the criterion for better prediction is one that forecasts a quantile closer to the quantile of the actual value, regardless of whether the prediction is closer in terms of the units of the outcome variable. In essence, the quantile transformation effectively compels the sign test to assess models based on their capacity to predict the quantile of the actual value, rather than their capability to predict the actual value itself.

\section{Results}

In what follows, the performance of \href{https://github.com/WojtekFulmyk/mlcausality}{\textbf{mlcausality}} (MLC in the following plots) is compared to the following three algorithms: the large-scale nonlinear granger causality (lsNGC) algorithm \citep{wismuller2021large}; the mutual nonlinear cross-mapping ­methods using local models (LM) algorithm (\cite{sugihara2012detecting}, using the software implementation provided by \cite{causalccmsoftware2021}); and the PC-momentary conditional independence (PCMCI) algorithm \citep{runge2019detecting}. For lsNGC, $c_f = 25$ and $c_g = 5$, which are suggested in \cite{wismuller2021large}, unless the lsNGC algorithm threw an error, at which point $c_f$ was progressively decreased until all simulated data ran successfully. For all other algorithms the software defaults were kept.

All models were run on a computer with an AMD Ryzen 5 3600 6-Core processor with 12 threads in total. In order to ensure that the comparison between \href{https://github.com/WojtekFulmyk/mlcausality}{\textbf{mlcausality}} and other competing algorithms is as fair as possible, all algorithms were parallelized to run on 12 processes, one for each thread.

The results presented herein are for lag orders selected using Cao's minimum embedding dimension selection method \citep{cao1997practical}. In particular, for every network and time-series length combination, Cao's algorithm is run on all time-series in that combination, with the largest identified minimum embedding dimension used as the basis for construction that combination's lag order.

\subsection{The Simulated Networks}

\begin{figure}[!htb]
\captionsetup{font=footnotesize}
\captionsetup{labelfont=bf}
\begin{subfigure}{0.32\textwidth}
  \includegraphics[width=\linewidth]{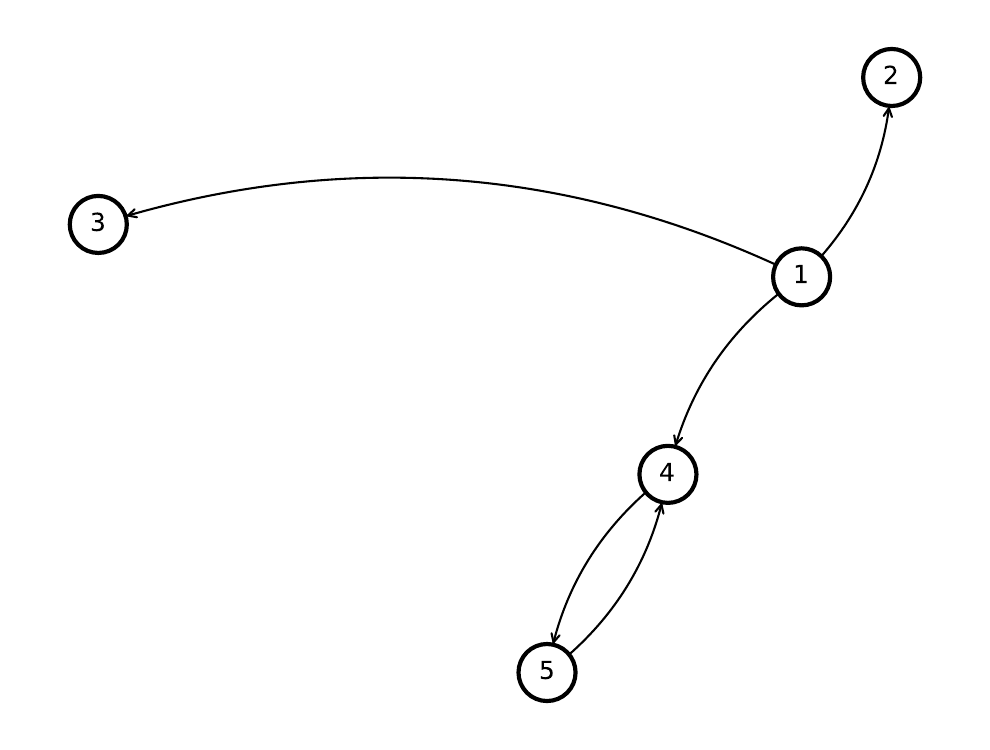}
  \caption{5 Linear and 5 Nonlinear}
\end{subfigure}
\hfill
\begin{subfigure}{0.32\textwidth}
  \includegraphics[width=\linewidth]{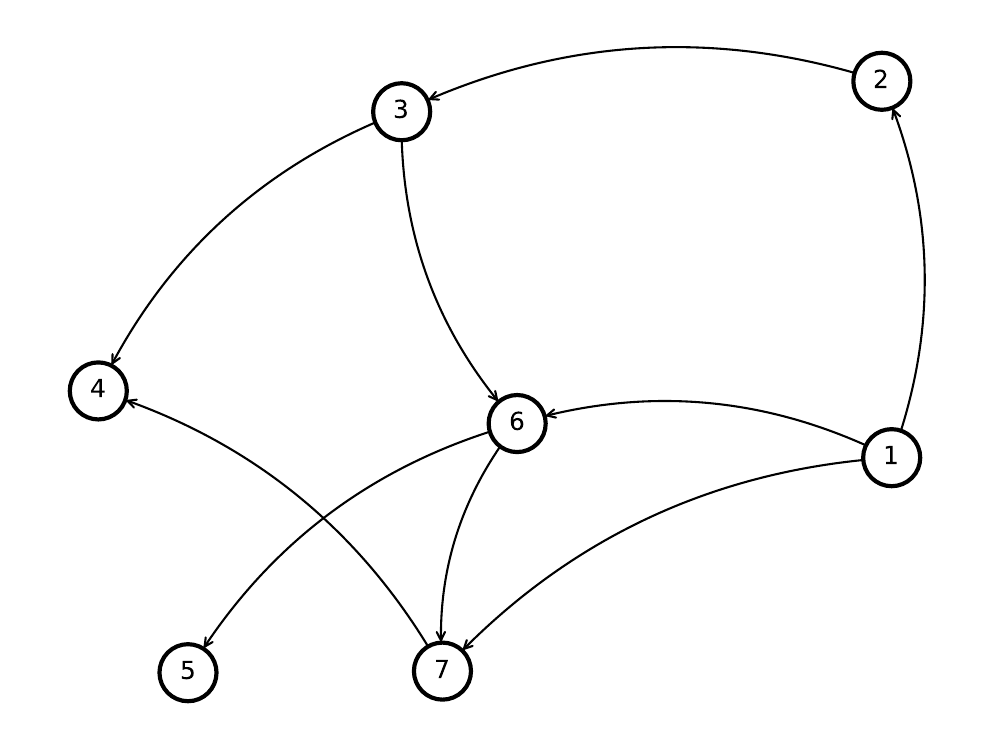}
  \caption{7 Nonlinear}
\end{subfigure}
\hfill
\begin{subfigure}{0.32\textwidth}%
  \includegraphics[width=\linewidth]{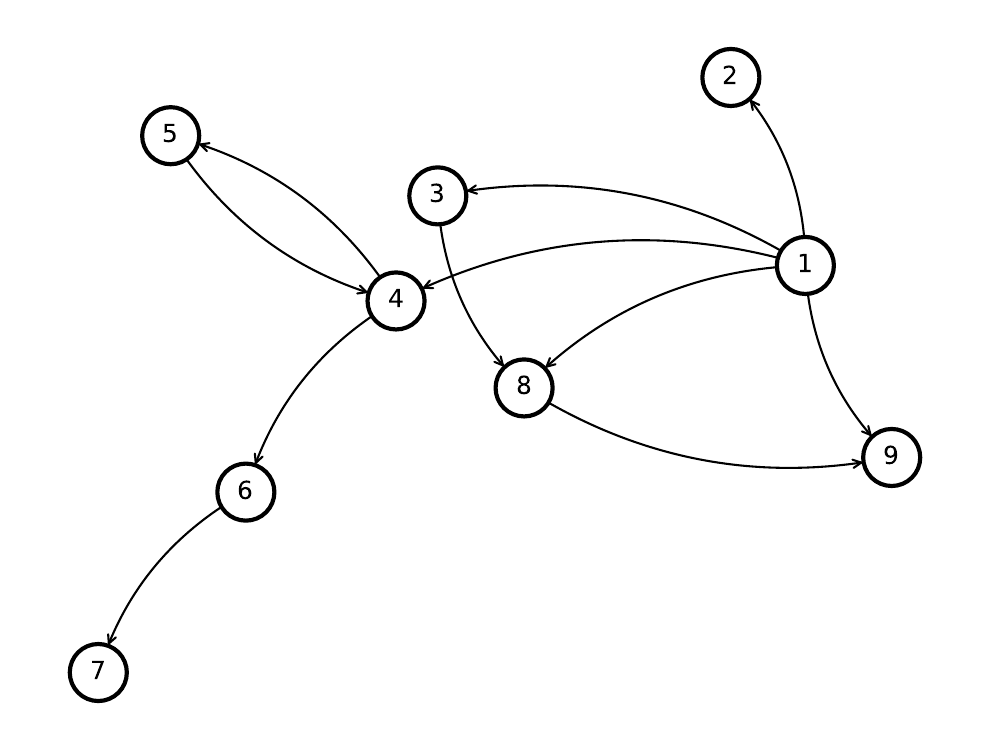}
  \caption{9 Nonlinear}
\end{subfigure}
\hfill
\begin{subfigure}{0.32\textwidth}%
  \includegraphics[width=\linewidth]{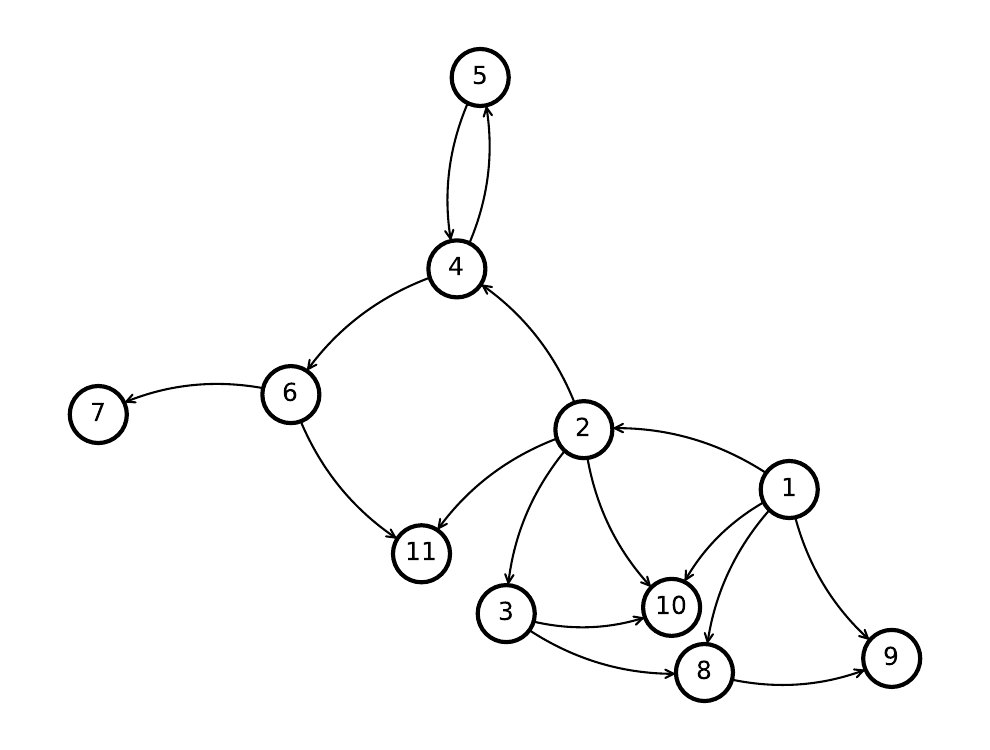}
  \caption{11 Nonlinear}
\end{subfigure}
\hfill
\begin{subfigure}{0.32\textwidth}%
  \includegraphics[width=\linewidth]{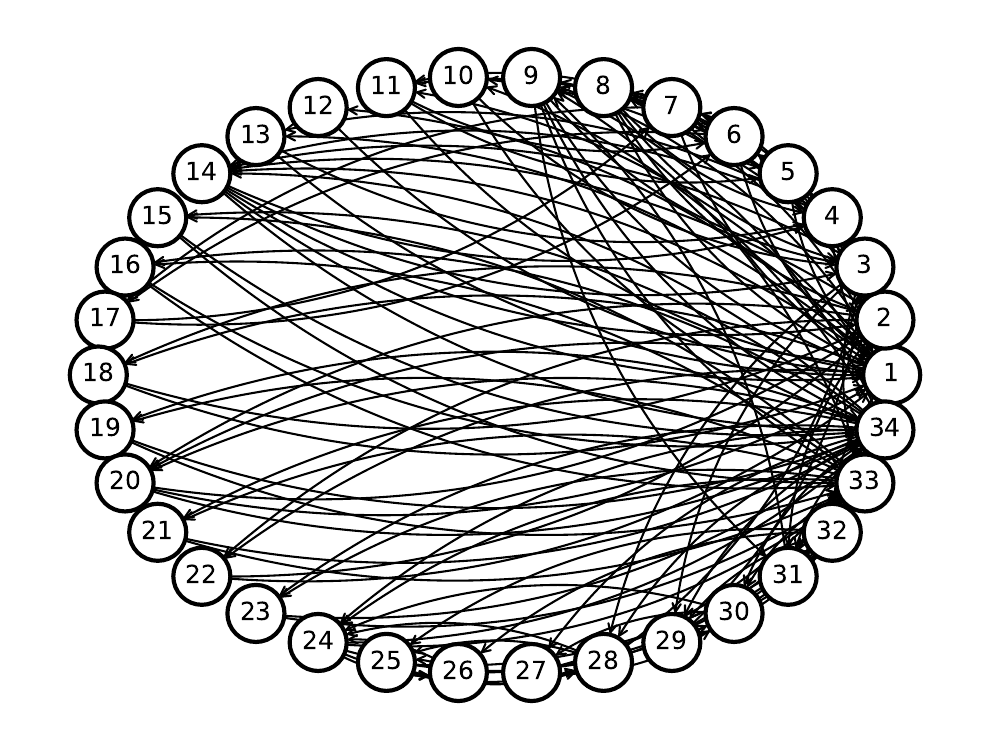}
  \caption{34 Zachary1}
\end{subfigure}
\hfill
\begin{subfigure}{0.32\textwidth}%
  \includegraphics[width=\linewidth]{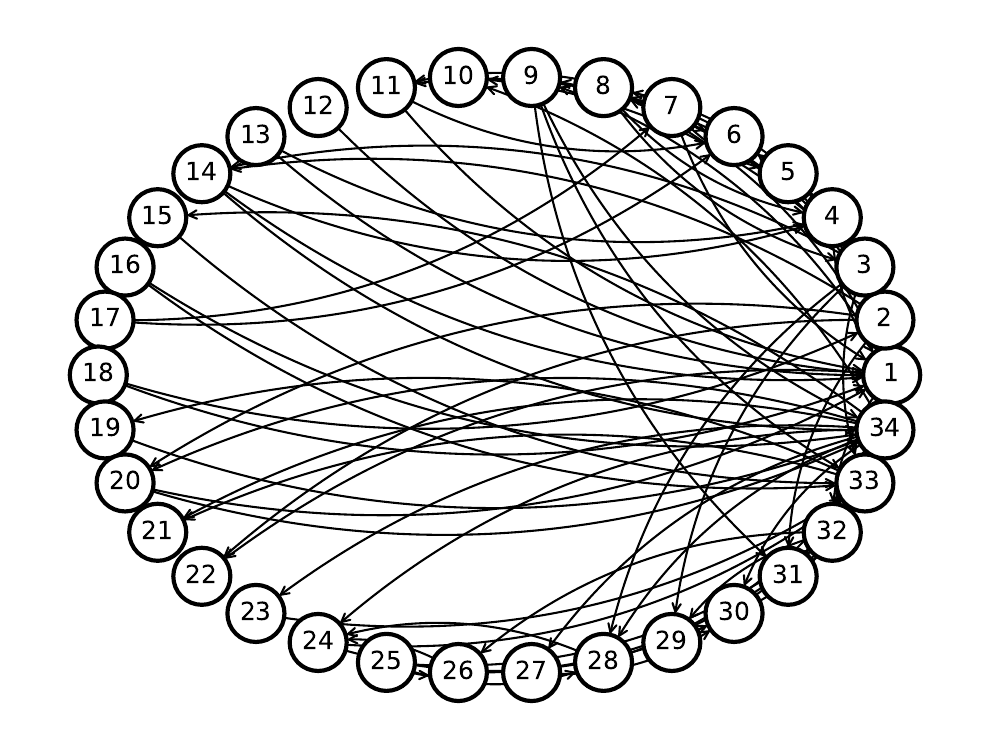}
  \caption{34 Zachary2}
\end{subfigure}
\captionsetup{justification=raggedright,singlelinecheck=false}
\caption{Network plots}\label{fig:network_plots}
\end{figure}

The following briefly describes the data generating processes for the simulated networks for which performance results are presented. All networks were initialized using normally distributed white noise $w(t)$ with mean = 0 and variance = 1. Every network was analyzed for time-series lengths of 500, 1000, 1500, and 2000 time-points. For every network and every time-series length, 50 independent sets were generated after discarding a 500 time-point burn-in. The network plots for all tested networks are available in figure \ref{fig:network_plots}.

\textit{5-node linear network:} This network was first proposed as example 3 in \cite{baccala2001partial}. The network is generated using the following multivariate autoregressive model:
\begin{equation}
{\normalsize \begin{aligned}
&x_1(t) = 0.95\sqrt{2}x_1(t-1) - 0.9025x_1(t-2) + w_1(t) \\
&x_2(t) = 0.5x_1(t-2) + w_2(t) \\
&x_3(t) = -0.4x_1(t-3) + w_3(t) \\
&x_4(t) = -0.5x_1(t-2) + 0.25\sqrt{2}x_4(t-1) + 0.25\sqrt{2}x_5(t-1) + w_4(t) \\
&x_5(t) = -0.25\sqrt{2}x_4(t-1) + 0.25\sqrt{2}x_5(t-1) + w_5(t) \\
\end{aligned}}
\end{equation}

\noindent This network has the following causal connections: $x_1 \rightarrow x_2$,  $x_1 \rightarrow x_3$, $x_1 \rightarrow x_4$, $x_4 \rightarrow x_5$, and $x_5 \rightarrow x_4$. This network structure has the potential to present several difficulties for a Granger causality recovery algorithm. For instance, although there is no direct causal link between $x_2$, $x_3$, and $x_4$, they may all be correlated because of the causal effect of $x_1$ on all of them.

\textit{5-node nonlinear network:} This network was introduced in \cite{wismuller2021large}. The causal connections in this 5-node nonlinear network are identical to those of the 5-node linear network, except some of those connections are converted into nonlinear ones:
\begin{equation}
{\normalsize \begin{aligned}
&x_1(t) = 0.95\sqrt{2}x_1(t-1) - 0.9025x_1(t-2) + w_1(t) \\
&x_2(t) = 0.5x_1^2(t-2) + w_2(t) \\
&x_3(t) = -0.4x_1(t-3) + w_3(t) \\
&x_4(t) = -0.5x_1^2(t-2) + 0.5\sqrt{2}x_4(t-1) + 0.25\sqrt{2}x_5(t-1) + w_4(t) \\
&x_5(t) = -0.5\sqrt{2}x_4(t-1) + 0.5\sqrt{2}x_5(t-1) + w_5(t) \\
\end{aligned}}
\end{equation}

\textit{7-node nonlinear network:} This network introduces many very complicated nonlinear interactions:
\begin{equation}
{\normalsize \begin{aligned}
&x_1(t) = 0.95\sqrt{2}x_1(t-1) - 0.9025x_1(t-2) + w_1(t) \\
&x_2(t) = -0.04x_1^3(t-3) + 0.04x_1^3(t-1) + w_2(t) \\
&x_3(t) = -0.04\sqrt{2}x_2^3(t-1) + 0.04\sqrt{2}x_2^3(t-2) + w_3(t) \\
&x_4(t) = ln(1+|x_3(t-1)|)*sgn(x_3(t-1)) + 0.001x_7^3(t-2) - 0.001x_7^3(t-3) + w_4(t) \\
&x_5(t) = 0.04*clip(w_{5a}(t),-1,1)*x_6^5(t-2) + w_{5b}(t) \\
&x_6(t) = 0.04*x_1^3(t-2) + 0.04*x_3^3(t-1) + w_6(t) \\
&x_7(t) = clip(w_{7a}(t),-0.5,0.5)*(0.04x_1^3(t-2) + 0.1x_6^2(t-1) - 0.1x_6^2(t-2)) + w_{7b}(t) \\
\end{aligned}}
\end{equation}

\noindent where $clip(v,a,b)$ is a function that limits $v$ to the range $[a,b]$ and $sgn(v)$ is the sign of $v$. The connections for this network are as follows: $x_1 \rightarrow x_2$,  $x_1 \rightarrow x_6$, $x_1 \rightarrow x_7$, $x_2 \rightarrow x_3$, $x_3 \rightarrow x_4$, $x_3 \rightarrow x_6$, $x_6 \rightarrow x_5$, $x_6 \rightarrow x_7$, and $x_7 \rightarrow x_4$.

\textit{9-node nonlinear network:} This network combines many nonlinear interactions with autoregressive terms for all variables:
\begin{equation}
{\normalsize \begin{aligned}
&x_1(t) = 0.95\sqrt{2}x_1(t-1) - 0.9025x_1(t-2) + w_1(t) \\
&x_2(t) = 0.5x_1^2(t-2) + 0.5x_2^2(t-1) - 0.4x_2^2(t-2) + w_2(t) \\
&x_3(t) = -0.4x_1(t-3) + 0.5x_3^2(t-1) - 0.4x_3^2(t-2) + w_3(t) \\
&x_4(t) = -0.5x_1^2(t-2) + 0.5x_4^2(t-1) - 0.4x_4^2(t-2) + 0.5\sqrt{2}x_4(t-1) + 0.25\sqrt{2}x_5(t-1) + w_4(t) \\
&x_5(t) = -0.5\sqrt{2}x_4(t-1) + 0.5\sqrt{2}x_5(t-1) + w_5(t) \\
&x_6(t) = sgn(x_4(t-1))*ln(|x_4(t-1)| + 1) + 0.5x_6^2(t-1) - 0.4x_6^2(t-2) + w_6(t) \\
&x_7(t) = 0.04*clip(w_{7a}(t),-1,1)*x_6^5(t-2) + 0.5x_7^2(t-1) - 0.4x_7^2(t-2) + w_{7b}(t) \\
&x_8(t) = 0.4x_1(t-2) + 0.25x_3^3(t-1) + 0.5x_8^2(t-1) - 0.4x_8^2(t-2) + w_8(t) \\
&x_9(t) = clip(w_{9a}(t),-0.5,0.5)*(0.2x_1(t-2) + 0.1x_8^2(t-1) - 0.1x_8^2(t-2)) + 0.5x_9^2(t-1) - 0.4x_9^2(t-2) + w_{9b}(t) \\
\end{aligned}}
\end{equation}

\noindent where $clip(v,a,b)$ is a function that limits $v$ to the range $[a,b]$ and $sgn(v)$ is the sign of $v$. The connections of this network are as follows: $x_1 \rightarrow x_2$,  $x_1 \rightarrow x_3$, $x_1 \rightarrow x_4$, $x_1 \rightarrow x_8$, $x_1 \rightarrow x_9$, $x_3 \rightarrow x_8$, $x_4 \rightarrow x_5$, $x_4 \rightarrow x_6$, $x_5 \rightarrow x_4$, $x_6 \rightarrow x_7$, and $x_8 \rightarrow x_9$.

\textit{11-node nonlinear network:} Yet another network with many nonlinear interactions:
\begin{equation}
{\normalsize \begin{aligned}
&x_1(t) = 0.25x_1^2(t-1) - 0.25x_1^2(t-2) + w_1(t) \\
&x_2(t) = ln(1+|x_1(t-2)|)*sgn(x_1(t-2)) + w_2(t) \\
&x_3(t) = -0.1x_2^3(t-3) + w_3(t) \\
&x_4(t) = -0.5x_2^2(t-2) + 0.5\sqrt{2}x_4(t-1) + 0.25\sqrt{2}x_5(t-1) + w_4(t) \\
&x_5(t) = -0.5\sqrt{2}x_4(t-1) + 0.5\sqrt{2}x_5(t-1) + w_5(t) \\
&x_6(t) = ln(1+|x_4(t-1)|)*sgn(x_4(t-1)) + w_6(t) \\
&x_7(t) = 0.04*clip(w_{7a}(t),-1,1)*x_6^5(t-2) + w_{7b}(t) \\
&x_8(t) = 0.4x_1(t-2) + 0.25x_3^3(t-1) + w_8(t) \\
&x_9(t) = clip(w_{9a}(t),-0.5,0.5)*(0.2x_1(t-2) + 0.1x_8^2(t-1) - 0.1x_8^2(t-2)) + w_{9b}(t) \\
&x_{10}(t) = 0.25x_1^2(t-3) - 0.01x_2^2(t-3) + 0.15x_3^3(t-3) + w_{10}(t) \\
&x_{11}(t) = 0.1x_2^4(t-1) - 0.1x_2^4(t-2) + 0.1x_6^3(t-3) + w_{11}(t) \\
\end{aligned}}
\end{equation}

\noindent where $clip(v,a,b)$ is a function that limits $v$ to the range $[a,b]$ and $sgn(v)$ is the sign of $v$. The connections of this network are as follows: $x_1 \rightarrow x_2$,  $x_1 \rightarrow x_8$, $x_1 \rightarrow x_9$, $x_1 \rightarrow x_{10}$, $x_2 \rightarrow x_3$, $x_2 \rightarrow x_4$, $x_2 \rightarrow x_{10}$, $x_2 \rightarrow x_{11}$, $x_3 \rightarrow x_8$, $x_3 \rightarrow x_{10}$, $x_4 \rightarrow x_{5}$, $x_4 \rightarrow x_{6}$, $x_5 \rightarrow x_{4}$, $x_6 \rightarrow x_{7}$, and $x_8 \rightarrow x_9$.

\textit{34-node Zachary1 and Zachary2 networks:} These networks are identical to the Zachary1 and Zachary2 networks in \cite{wismuller2021large} and are constructed using the undirected connections in the Zachary karate club dataset \citep{zachary1977information}. The Zachary dataset is a social network composed of 34 members of a karate club that lists links between pairs of members that interacted outside the club. The nodal interactions, adapted from \cite{marinazzo2008kernel}, are as follows:
\begin{equation}
{\normalsize \begin{aligned}
&x_i(t) = \bigg(1 - \sum_{j=1}^n c_{ij}\bigg)(1-ax_{i}^2(t-1)) + \sum_{j=1}^n c_{ij} (1-ax_{j}^2(t-1)) + sw_{i}(t)
\end{aligned}}
\end{equation}

\noindent where $c_{ij}$ indicates the coupling $j \rightarrow i$. For the Zachary1 network, all 78 edges linking the 34 nodes in the network are assumed to be bidirectional and $a = 1.8$, $s= 0.01$, and $c=0.025$. For the Zachary2 network, 5 of the 78 edges linking the 34 nodes in the network are randomly selected to be bidirectional, while for the rest of the links, the direction is assigned randomly. Moreover, for the Zachary2 network $c=0.05$. Note that each of the 50 independent sets of Zachary2 networks could have a different network structure because the directions of the links are randomly assigned for each of the 50 independent sets.

\subsection{Evaluating mlcausality's Network Recovery Performance}

\subsubsection{Non-thresholded Metrics}

\begin{figure}[!htb]
\captionsetup{font=footnotesize}
\captionsetup{labelfont=bf}
\captionsetup{justification=raggedright,singlelinecheck=false}
\includegraphics[width=\linewidth]{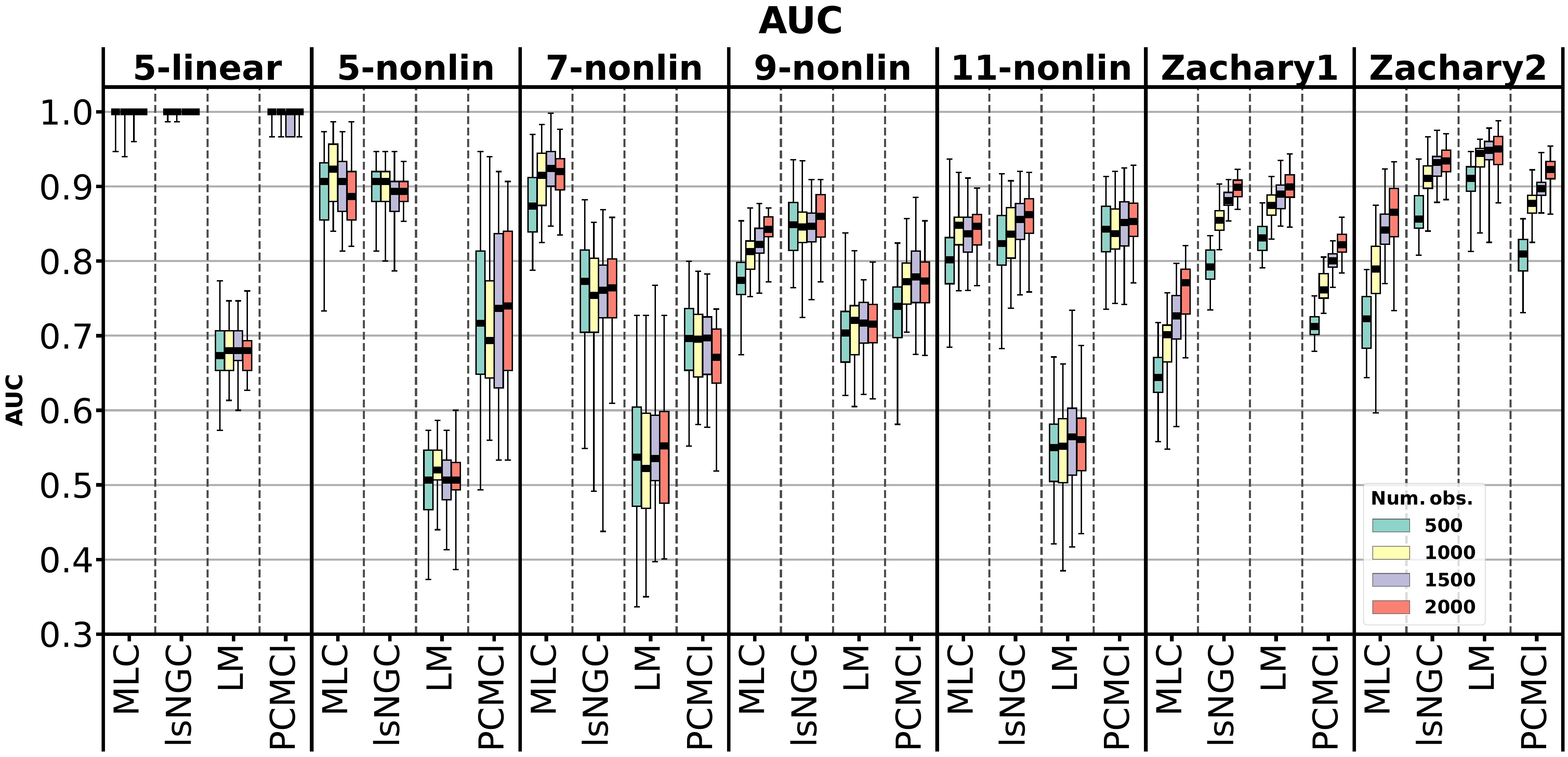}
\caption{AUC boxplots for time-series of different lengths. In all cases the median of the distribution is represented by a black square; the upper and lower edges of a box represent the 25th and 75th percentiles; and the whiskers extend to the distribution maximums and minimums. MLC indicates the \textbf{mlcaulsality} model; lsNGC indicates the large-scale nonlinear granger causality algorithm; LM indicates the mutual nonlinear cross-mapping ­methods using local models approach; and PCMCI indicates the PC-momentary conditional independence algorithm.}\label{fig:aucplotmulti}
\end{figure}

Figure \ref{fig:aucplotmulti} shows the area under the receiver operating characteristic curve (AUC) for all model, network, and time-series length combinations. For nonlinear networks up to around 11 nodes in size \href{https://github.com/WojtekFulmyk/mlcausality}{\textbf{mlcausality}} with kernel ridge regression exhibits leading, joint-leading, or near-leading AUC performance, with AUC performance significantly declining for the 34-node Zachary1 and Zachary2 networks. For networks with 9 nodes or less substantial performance improvements are not observed as the time-series length increases. For the 11-node nonlinear, Zachary1, and Zachary2 networks \href{https://github.com/WojtekFulmyk/mlcausality}{\textbf{mlcausality}}'s performance steadily improves as the time-series length increases: this indicates that for networks greater than around 10 nodes in length at least 2000 time-points are needed to achieve peak performance with the default 70-30 train-test split. A similar pattern is observed for rival algorithms when recovering the Zachary1 and Zachary2 networks but not the 11-node nonlinear network. This implies that \href{https://github.com/WojtekFulmyk/mlcausality}{\textbf{mlcausality}} is more ``data hungry'' than competing algorithms, a fact that should not be surprising given that \href{https://github.com/WojtekFulmyk/mlcausality}{\textbf{mlcausality}} splits the data into separate train and test sets. It can therefore be concluded that for networks of up to 10 nodes in size that have at least 500 time-points \href{https://github.com/WojtekFulmyk/mlcausality}{\textbf{mlcausality}} achieves leading or near-leading AUC performance. Moreover, given the data-hungry nature of \href{https://github.com/WojtekFulmyk/mlcausality}{\textbf{mlcausality}} and the trajectory of improving AUC scores as the number of time-points increases for the 34-node Zachary1 and Zachary2 networks, it is plausible to assume that \href{https://github.com/WojtekFulmyk/mlcausality}{\textbf{mlcausality}} might be capable of matching or exceeding the AUC performance of other algorithms given sufficently long time-series data.

\begin{figure}[!htb]
\captionsetup{font=footnotesize}
\captionsetup{labelfont=bf}
\captionsetup{justification=raggedright,singlelinecheck=false}
\includegraphics[width=\linewidth]{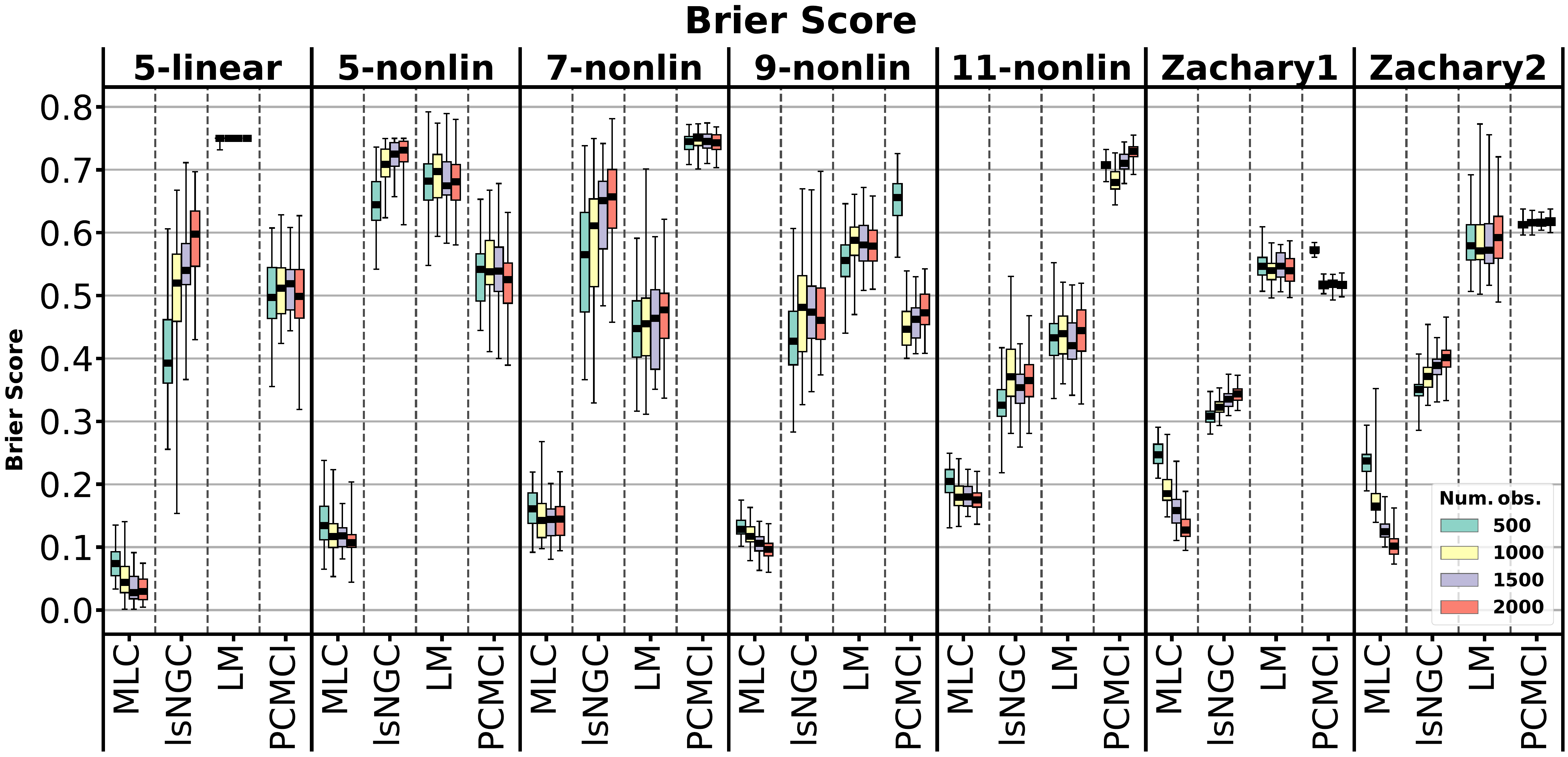}
\caption{Brier score boxplots for time-series of different lengths. In all cases the median of the distribution is represented by a black square; the upper and lower edges of a box represent the 25th and 75th percentiles; and the whiskers extend to the distribution maximums and minimums. MLC indicates the \textbf{mlcaulsality} model; lsNGC indicates the large-scale nonlinear granger causality algorithm; LM indicates the mutual nonlinear cross-mapping ­methods using local models approach; and PCMCI indicates the PC-momentary conditional independence algorithm.}\label{fig:brierplotmulti}
\end{figure}

Figure \ref{fig:brierplotmulti} compares the Brier scores for all network, time-series length, and model combinations. Here, we see that \href{https://github.com/WojtekFulmyk/mlcausality}{\textbf{mlcausality}} exhibits significantly lower Brier scores than competing algorithms. The implication of lower Brier scores is that \href{https://github.com/WojtekFulmyk/mlcausality}{\textbf{mlcausality}}'s $p$-values are better calibrated than rival algorithms and are a truer reflection of actual probabilities. Furthermore, for most of the analyzed networks \href{https://github.com/WojtekFulmyk/mlcausality}{\textbf{mlcausality}}'s Brier scores tend to decrease as the time-series length increases: this implies that the $p$-values tend to converge towards true probabilities as time-series lengths increase. The same is not true for any of the other tested networks, with lsNGC in particular generally exhibiting higher Brier scores with increased time-series lengths. The above suggests that $p$-value-based thresholding rules are likely to perform better for \href{https://github.com/WojtekFulmyk/mlcausality}{\textbf{mlcausality}} than competing algorithms, especially with long time-series data. 

\subsubsection{Metrics at G-mean Optimal Thresholds}

\begin{figure}[!htb]
\captionsetup{font=footnotesize}
\captionsetup{labelfont=bf}
\captionsetup{justification=raggedright,singlelinecheck=false}
\includegraphics[width=\linewidth]{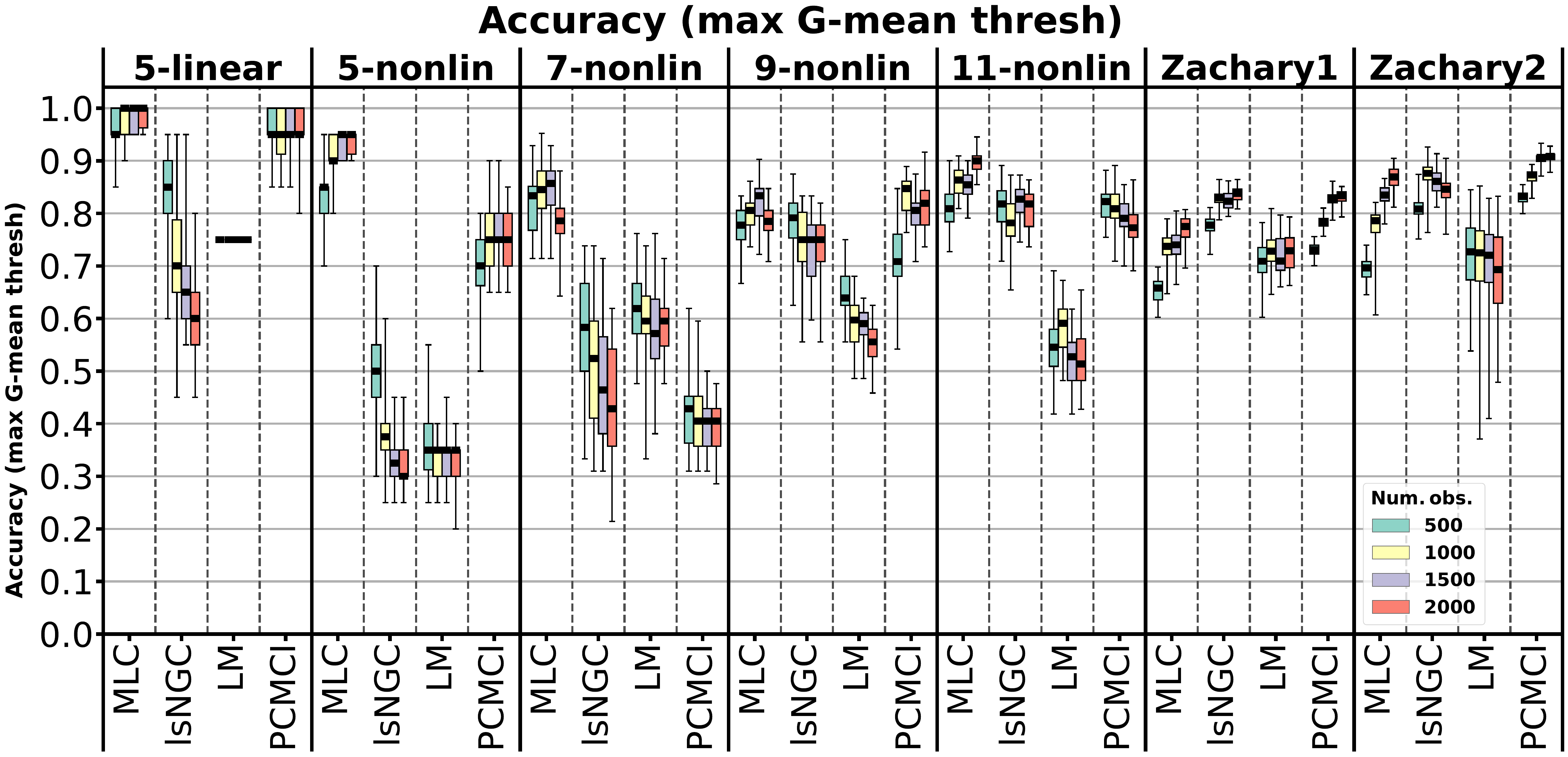}
\caption{Accuracy boxplots for a threshold at the G-mean max for time-series of different lengths. In all cases the median of the distribution is represented by a black square; the upper and lower edges of a box represent the 25th and 75th percentiles; and the whiskers extend to the distribution maximums and minimums. MLC indicates the \textbf{mlcaulsality} model; lsNGC indicates the large-scale nonlinear granger causality algorithm; LM indicates the mutual nonlinear cross-mapping ­methods using local models approach; and PCMCI indicates the PC-momentary conditional independence algorithm.}\label{fig:accuracy2plotmulti}
\end{figure}

Figure \ref{fig:accuracy2plotmulti} shows the accuracy at optimal geometric mean of sensitivity and specificity (G-mean) $p$-value thresholds. Specifically, for every model, network, and time-series combination, 100 equally-spaced $p$-value thresholds were tested, with the $p$-value threshold that generated the highest median G-mean for the 50 independent sets in that combination being chosen. Figure \ref{fig:accuracy2plotmulti} indicates that \href{https://github.com/WojtekFulmyk/mlcausality}{\textbf{mlcausality}} exhibits leading or joint-leading accuracy at optimal $p$-value thresholds for networks up to and including 11 nodes in size, and near-leading accuracy for the 34-node Zachary networks when the number of observations is high (2000+ time-points).

Moreover, figure \ref{fig:accuracy2plotmulti} suggests that relying on AUC scores alone can be a somewhat misleading indicator of true model performance when a thresholding criteria must be implemented. For instance, despite near-perfect AUC scores for the lsNGC model on the 5-node linear network, at optimal G-mean thresholds, the accuracies for those same model and network combinations are well below 1. This seemingly disturbing discrepancy is not the result of an error. The lsNGC AUC scores indicate that, for nearly every independent instance of the 5-node linear model, there exists some split that leads to perfect classification. On the other hand, the accuracy scores in figure \ref{fig:accuracy2plotmulti} are evaluated at thresholds that maximize the median G-mean for \emph{all} 50 independent sets in a network--time-series combination. Although almost every single instance of the 5-node linear network can be perfectly classified by some $p$-value split when using the lsNGC model, that split is not stable from one independent set of data to the next. Furthermore, as the number of time-points increases, the accuracy of the lsNGC model tends to decline, which is entirely consistent with the increasing Brier scores in time-series lengths observed in figure \ref{fig:brierplotmulti}.

\begin{figure}[!htb]
\captionsetup{font=footnotesize}
\captionsetup{labelfont=bf}
\captionsetup{justification=raggedright,singlelinecheck=false}
\includegraphics[width=\linewidth]{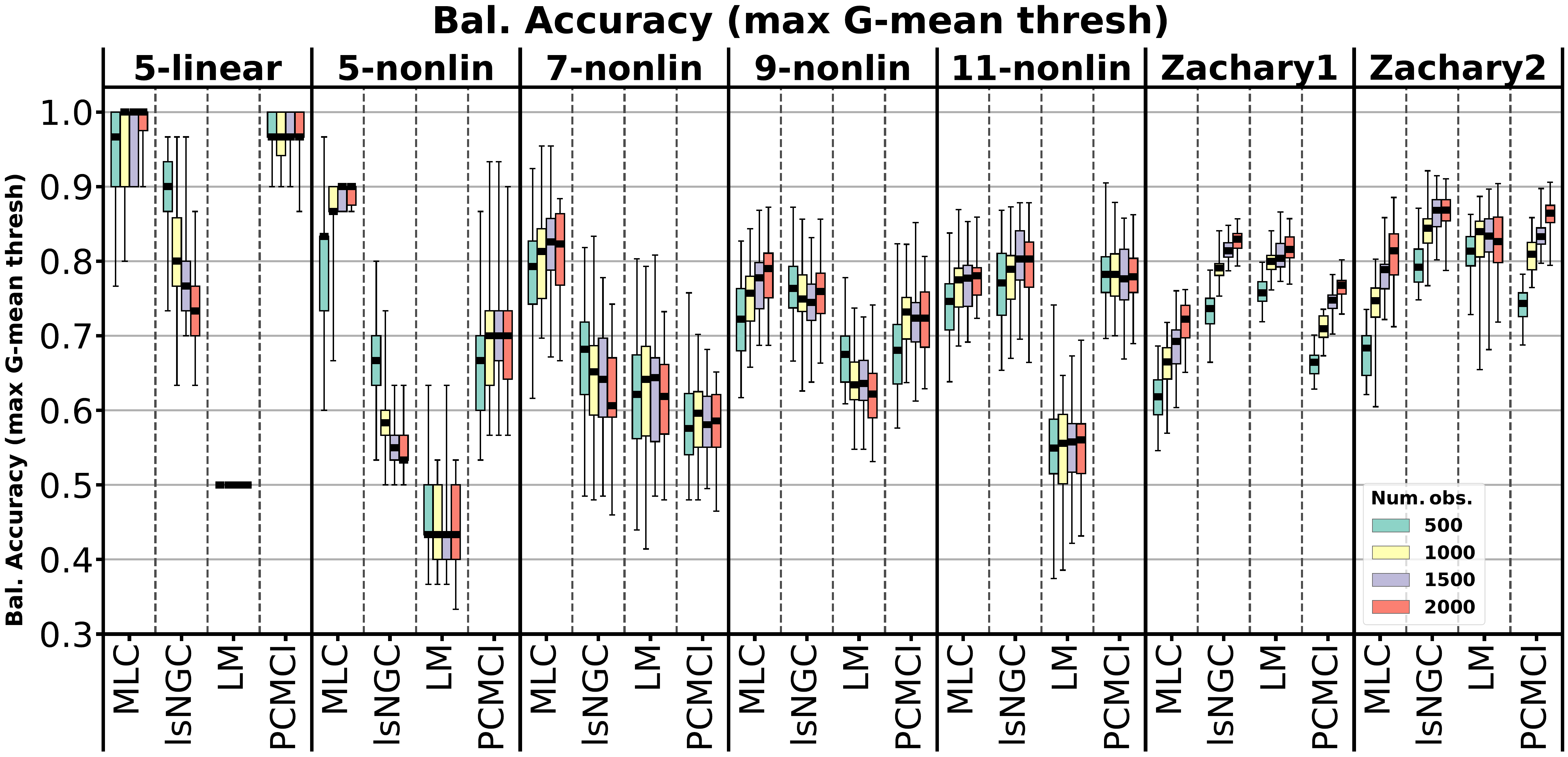}
\caption{Balanced accuracy boxplots for a threshold at the G-mean max for time-series of different lengths. In all cases the median of the distribution is represented by a black square; the upper and lower edges of a box represent the 25th and 75th percentiles; and the whiskers extend to the distribution maximums and minimums. MLC indicates the \textbf{mlcaulsality} model; lsNGC indicates the large-scale nonlinear granger causality algorithm; LM indicates the mutual nonlinear cross-mapping ­methods using local models approach; and PCMCI indicates the PC-momentary conditional independence algorithm.}\label{fig:balancedaccuracy2plotmulti}
\end{figure}

Figure \ref{fig:balancedaccuracy2plotmulti} shows the balanced accuracy scores at the maximal G-mean threshold. Balanced accuracy differs from accuracy in that balanced accuracy measures the average of accuracy scores for both the minority and majority classes (in effect, this is equivalent to taking the average of the sensitivity and the specificity). The balanced accuracy scores confirm that \textbf{mlcaulsality} exhibits leading performance at optimal $p$-value thresholds for networks up to and including 9 nodes in size. For the 11-node and Zachary2 networks \textbf{mlcaulsality} achieves near-leading balanced accuracy only when the number of observations is very high (2000+ time-points). For the Zachary1 network near-leading balanced accuracy is never achieved, but \textbf{mlcaulsality} does exhibit rapidly rising balanced accuracy scores in time-series length which suggests that leading or near-leading performance may be possible with a sufficiently long time-series data.

\subsubsection{Metrics at the $p$-value = 0.05 Threshold}

When Granger causality algorithms are used in practice, the true nature of the Granger causal relationships in the studied networks are unknown to the researcher, and therefore the researcher will not be able to use an optimal threshold derived from unobserved sensitivity and specificity such as G-mean or Youden's index. However, the researcher will have to pick and settle on some threshold anyway in order to recover the relationships and make the analysis meaningful. Figures \ref{fig:accuracyplotmulti} and  \ref{fig:balancedaccuracyplotmulti} below show the accuracy and balanced accuracy scores when the $p$-value threshold is set to 0.05, a commonly used significance level. Comparing accuracy and balanced accuracy scores in figures \ref{fig:accuracyplotmulti} and  \ref{fig:balancedaccuracyplotmulti} to their counterparts at G-mean optimal thresholds in figures \ref{fig:accuracy2plotmulti} and \ref{fig:balancedaccuracy2plotmulti} reveals that the 0.05 $p$-value threshold yields excellent, near-optimal results for the \href{https://github.com/WojtekFulmyk/mlcausality}{\textbf{mlcausality}} algorithm for the analyzed networks. Moreover, at the 0.05 threshold, \href{https://github.com/WojtekFulmyk/mlcausality}{\textbf{mlcausality}} achieves superior accuracy to all competing networks, and leading or joint-leading balanced accuracy scores for networks of 9 nodes or less. For networks with 11 nodes or more \href{https://github.com/WojtekFulmyk/mlcausality}{\textbf{mlcausality}}'s balanced accuracy at the 0.05 threshold increases with time-series length, which suggests that, for sufficiently long time-series, \href{https://github.com/WojtekFulmyk/mlcausality}{\textbf{mlcausality}} may be able to match or exceed the balanced accuracy performance of competing algorithms.

\begin{figure}[!htb]
\captionsetup{font=footnotesize}
\captionsetup{labelfont=bf}
\captionsetup{justification=raggedright,singlelinecheck=false}
\includegraphics[width=\linewidth]{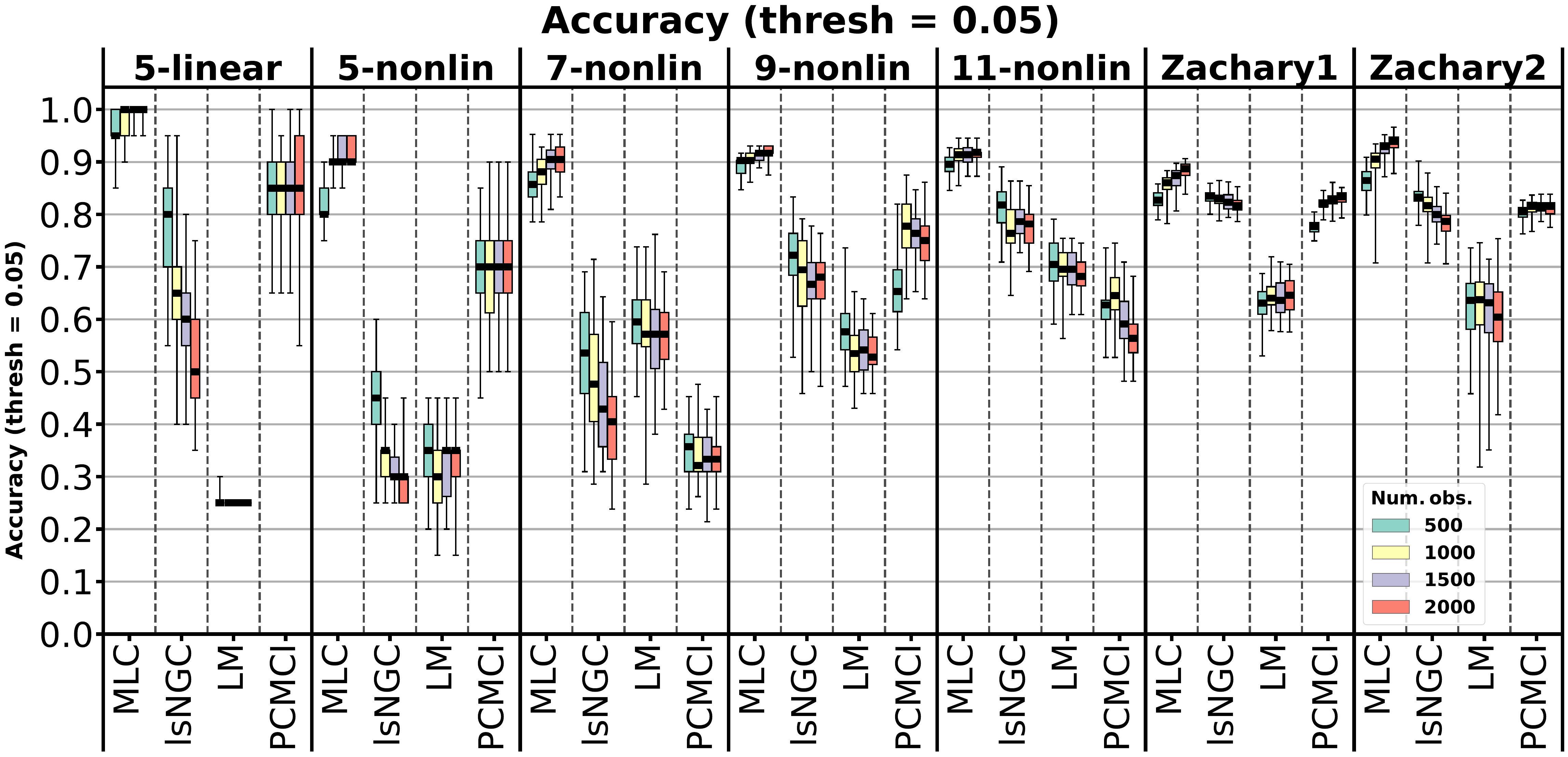}
\caption{Accuracy boxplots for a threshold at $p$-value=0.05 for time-series of different lengths. In all cases the median of the distribution is represented by a black square; the upper and lower edges of a box represent the 25th and 75th percentiles; and the whiskers extend to the distribution maximums and minimums. MLC indicates the \textbf{mlcaulsality} model; lsNGC indicates the large-scale nonlinear granger causality algorithm; LM indicates the mutual nonlinear cross-mapping ­methods using local models approach; and PCMCI indicates the PC-momentary conditional independence algorithm.}\label{fig:accuracyplotmulti}
\end{figure}

\begin{figure}[!htb]
\captionsetup{font=footnotesize}
\captionsetup{labelfont=bf}
\captionsetup{justification=raggedright,singlelinecheck=false}
\includegraphics[width=\linewidth]{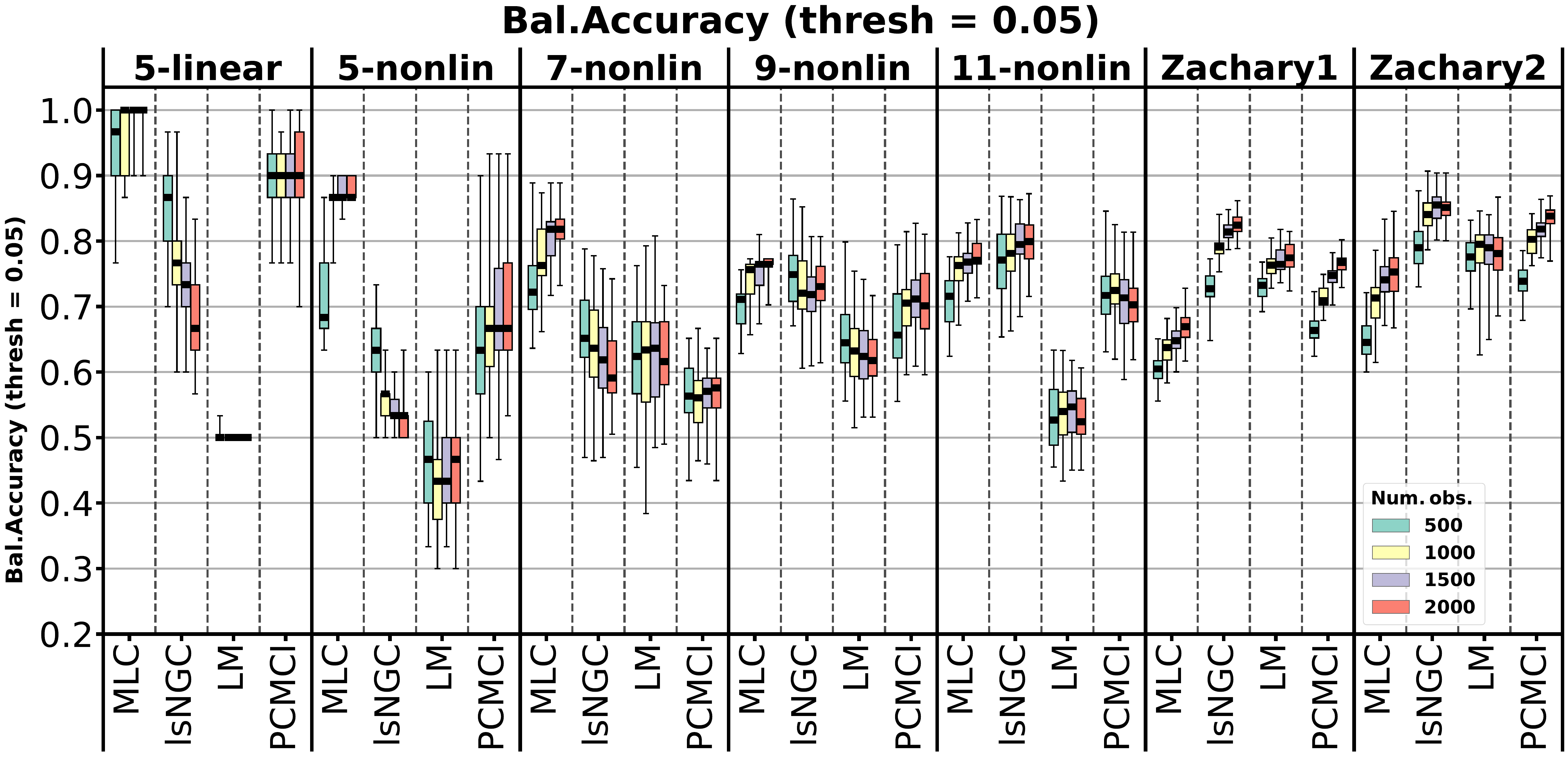}
\caption{Balanced accuracy boxplots for a threshold at $p$-value=0.05 for time-series of different lengths. In all cases the median of the distribution is represented by a black square; the upper and lower edges of a box represent the 25th and 75th percentiles; and the whiskers extend to the distribution maximums and minimums. MLC indicates the \textbf{mlcaulsality} model; lsNGC indicates the large-scale nonlinear granger causality algorithm; LM indicates the mutual nonlinear cross-mapping ­methods using local models approach; and PCMCI indicates the PC-momentary conditional independence algorithm.}\label{fig:balancedaccuracyplotmulti}
\end{figure}

\subsection{Evaluating mlcausality's Runtime Performance}

Table \ref{runtimes} presents the runtimes for all model, network, and time-series length combinations. In all cases, \href{https://github.com/WojtekFulmyk/mlcausality}{\textbf{mlcausality}} with kernel ridge regression and the RBF kernel runs significantly faster than rival algorithms, in some cases more than 10 times faster than the second fastest algorithm. Moreover, \href{https://github.com/WojtekFulmyk/mlcausality}{\textbf{mlcausality}} appears to scale much better, in terms of runtime, to increasing network sizes and time-series lengths than competing algorithms. When coupled with improving AUC, accuracy, and balanced accuracy performance in time-series length, \href{https://github.com/WojtekFulmyk/mlcausality}{\textbf{mlcausality}}'s excellent runtime performance provides strong arguments in favor of its usage to handle exceptionally large and complex time-series networks.

\begin{table}[h]
\footnotesize
\captionsetup{font=footnotesize}
\captionsetup{labelfont=bf}
\captionsetup{justification=raggedright,singlelinecheck=false}
\centering
\begin{tabularx}{\textwidth}{|cc||YYYY|} \hline
\multicolumn{2}{|c||}{\textbf{Network Details}} & \multicolumn{4}{c|}{\textbf{Runtime for 50 iterations using multiprocessing (seconds)}}\\ 
\textbf{Network} & \multicolumn{1}{c||}{\textbf{Num. obs.}} & \multicolumn{1}{c}{\href{https://github.com/WojtekFulmyk/mlcausality}{\textbf{mlcausality}}} & \multicolumn{1}{c}{\textbf{lsNGC}} & \multicolumn{1}{c}{\textbf{LM}} & \multicolumn{1}{c|}{\textbf{PCMCI}}\\ \toprule \hline
\multirow{4}{*}{\begin{tabular}{c}\textit{5-linear}\end{tabular}} & \textit{500} & <1 & 23 & 4 & 5 \\
& \textit{1000} & <1 & 47 & 11 & 6 \\
& \textit{1500} & 2 & 72 & 22 & 7 \\
& \textit{2000} & 5 & 95 & 36 & 7 \\ \hline
\multirow{4}{*}{\begin{tabular}{c}\textit{5-nonlinear}\end{tabular}} & \textit{500} & <1 & 24 & 4 & 4 \\
& \textit{1000} & <1 & 48 & 11 & 5 \\
& \textit{1500} & 2 & 72 & 22 & 6 \\
& \textit{2000} & 5 & 95 & 35 & 6 \\ \hline
\multirow{4}{*}{\begin{tabular}{c}\textit{7-nonlinear}\end{tabular}} & \textit{500} & <1 & 36 & 9 & 9 \\
& \textit{1000} & 1 & 73 & 24 & 11 \\
& \textit{1500} & 3 & 111 & 45 & 14 \\
& \textit{2000} & 7 & 147 & 72 & 17 \\ \hline
\multirow{4}{*}{\begin{tabular}{c}\textit{9-nonlinear}\end{tabular}} & \textit{500} & <1 & 60 & 16 & 11 \\
& \textit{1000} & 1 & 120 & 42 & 13 \\
& \textit{1500} & 4 & 180 & 77 & 15 \\
& \textit{2000} & 9 & 237 & 120 & 17 \\ \hline
\multirow{4}{*}{\begin{tabular}{c}\textit{11-nonlinear}\end{tabular}} & \textit{500} & <1 & 74 & 25 & 17 \\
& \textit{1000} & 2 & 147 & 67 & 20 \\
& \textit{1500} & 6 & 220 & 124 & 23 \\
& \textit{2000} & 11 & 294 & 198 & 27 \\ \hline
\multirow{4}{*}{\begin{tabular}{c}\textit{34-Zachary1}\end{tabular}} & \textit{500} & 5 & 245 & 261 & 344 \\
& \textit{1000} & 15 & 479 & 674 & 319 \\
& \textit{1500} & 33 & 719 & 1262 & 403 \\
& \textit{2000} & 62 & 942 & 2012 & 473 \\ \hline
\multirow{4}{*}{\begin{tabular}{c}\textit{34-Zachary2}\end{tabular}} & \textit{500} & 5 & \textit{245} & 259 & 329 \\
& \textit{1000} & 16 & 492 & 680 & 438 \\
& \textit{1500} & 33 & 730 & 1267 & 532 \\
& \textit{2000} & 61 & 976 & 2006 & 632 \\ \hline
\end{tabularx}
\caption{Runtimes (in seconds) for different models and networks for a computer with an AMD Ryzen 5 3600 6-Core dual-thread processor and 32 Gb of ram. 50 independent sets of each network were evaluated. All algorithms were parallelized to run 12 processes, one for each thread on the Ryzen processor.}\label{runtimes}
\end{table}

\section{Conclusion}

In this paper I presented a new method and associated Python library, \href{https://github.com/WojtekFulmyk/mlcausality}{\textbf{mlcausality}}, for identifying nonlinear Granger causal relationships. Despite \href{https://github.com/WojtekFulmyk/mlcausality}{\textbf{mlcausality}} containing a plug-in architecture that allows for the usage of any nonlinear regressor, the analysis presented herein focused specifically on the kernel ridge regressor with the radial basis function kernel. On simulated networks \href{https://github.com/WojtekFulmyk/mlcausality}{\textbf{mlcausality}} with the kernel ridge regressor and the RBF kernel achieves excellent performance at runtimes that are, in many cases, up to 10 times lower than the second fastest competing algorithm. For moderately sized networks of up to 11 nodes \href{https://github.com/WojtekFulmyk/mlcausality}{\textbf{mlcausality}} achieves leading or highly competitive recovery performance as measured by AUC, and tends to achieve superior accuracy and balanced accuracy when thresholded using $p$-value-based criteria. Furthermore, \href{https://github.com/WojtekFulmyk/mlcausality}{\textbf{mlcausality}}'s improving AUC, accuracy, balanced accuracy and Brier scores in time-series length for large networks, when coupled with \href{https://github.com/WojtekFulmyk/mlcausality}{\textbf{mlcausality}}'s excellent runtime performance, position this algorithm well to handle large networks with many time-steps. In short, \href{https://github.com/WojtekFulmyk/mlcausality}{\textbf{mlcausality}} achieves leading nonlinear Granger causality performance at a fraction of the computational time of rival algorithms.

\section{Code Availability}

The \href{https://github.com/WojtekFulmyk/mlcausality}{\textbf{mlcausality}} library itself is publicly available at \url{https://github.com/WojtekFulmyk/mlcausality}. Replication codes for all results in this paper are available at \url{https://github.com/WojtekFulmyk/mlcausality-krr-paper-replication}.

\bibliography{mlcausalitykrr}

\end{document}